\documentclass[10pt,twocolumn,letterpaper]{article}

\usepackage{cvpr}              

\usepackage{amsmath,amsfonts,bm}

\def\eqref#1{equation~\ref{#1}}

\def\1{\bm{1}}

\def\rx{{\textnormal{x}}}
\def\ry{{\textnormal{y}}}
\def\rz{{\textnormal{z}}}

\def\rvv{{\mathbf{v}}}

\def\rvx{{\mathbf{x}}}

\def\vmu{{\bm{\mu}}}
\def\vtheta{{\bm{\theta}}}

\def\vv{{\bm{v}}}

\def\vx{{\bm{x}}}

\DeclareMathAlphabet{\mathsfit}{\encodingdefault}{\sfdefault}{m}{sl}
\SetMathAlphabet{\mathsfit}{bold}{\encodingdefault}{\sfdefault}{bx}{n}

\def\sC{{\mathbb{C}}}
\def\sD{{\mathbb{D}}}

\def\sM{{\mathbb{M}}}

\def\sX{{\mathbb{X}}}

\def\sZ{{\mathbb{Z}}}

\newcommand{\E}{\mathbb{E}}

\newcommand{\normltwo}{L^2}

\usepackage{graphicx}
\usepackage{amsmath}
\usepackage{amssymb}
\usepackage{booktabs}

\usepackage{url}
\usepackage{multirow}
\usepackage{booktabs}
\usepackage{pifont}
\usepackage{graphicx}
\usepackage{subcaption}
\usepackage[toc,page]{appendix}
\usepackage[accsupp]{axessibility} 
\newcommand{\xmark}{\ding{55}}

\usepackage[pagebackref,breaklinks,colorlinks]{hyperref}
\usepackage{stmaryrd}

\usepackage[capitalize]{cleveref}
\crefname{section}{Sec.}{Secs.}
\Crefname{section}{Section}{Sections}
\Crefname{table}{Table}{Tables}
\crefname{table}{Tab.}{Tabs.}

\def\cvprPaperID{10135} 
\def\confName{CVPR}
\def\confYear{2022}

\begin{document}

\title{General Incremental Learning with Domain-aware Categorical Representations}

\author{Jiangwei Xie\textsuperscript{\rm 1}\thanks{Both authors contributed equally. This work was supported by Shanghai Science and Technology Program 21010502700.}
	\quad Shipeng Yan\textsuperscript{\rm 1,3,4}$^{*}$
	\quad Xuming He\textsuperscript{\rm 1,2}\\
	\textsuperscript{\rm 1}School of Information Science and Technology, ShanghaiTech University \quad \\
	\textsuperscript{\rm 2}Shanghai Engineering Research Center of Intelligent Vision and Imaging\\
	\textsuperscript{\rm 3}Shanghai Institute of Microsystem and Information Technology,
	Chinese Academy of Sciences\\
	\textsuperscript{\rm 4}University of Chinese Academy of Sciences\\
	\{xiejw,yanshp,hexm\}@shanghaitech.edu.cn
}
\maketitle

\begin{abstract}
    Continual learning is an important problem for achieving human-level intelligence in real-world applications as an agent must continuously accumulate knowledge in response to streaming data/tasks. In this work, we consider a general and yet under-explored incremental learning problem in which both the class distribution and class-specific domain distribution change over time.
    In addition to the typical challenges in class incremental learning, this setting also faces the intra-class stability-plasticity dilemma and intra-class domain imbalance problems.
    To address above issues, we develop a novel domain-aware continual learning method based on the EM framework.
    Specifically, we introduce a flexible class representation based on the von Mises-Fisher mixture model to capture the intra-class structure, using an expansion-and-reduction strategy to dynamically increase the number of components according to the class complexity. Moreover, we design a bi-level balanced memory to cope with data imbalances within and across classes, which combines with a distillation loss to achieve better inter- and intra-class stability-plasticity trade-off.
    We conduct exhaustive experiments on three benchmarks: iDigits, iDomainNet and iCIFAR-20.
    The results show that our approach consistently outperforms previous methods by a significant margin, demonstrating its superiority.
\end{abstract}

\section{Introduction}\label{sec:intro}
In order to achieve human-level intelligence, it is indispensable for a learning system to continuously accumulate knowledge over time in an ever-changing environment, known as continual or incremental learning \cite{chalup@biological}.
To cope with real-world scenarios, we consider a \textit{general incremental learning} problem~\cite{LomonacoM17, BuzzegaBPAC20}, where both the class distribution and class-specific domain distributions of the incoming data continuously change across sequential learning sessions.
This requires a model to incrementally learn not only novel class concepts but also new variants of previously-learned concepts.

Existing works on the general incremental learning typically focuse on the online class incremental learning setting~\cite{LomonacoM17, BuzzegaBPAC20}, which has to sacrifice the performance due to its strict computation/memory constraints. In this work, we instead aim to tackle the offline incremental learning setting, which has the potential to achieve significantly higher performance than the online counterpart. 
We note that, unlike the offline class incremental learning~\cite{delange2021continual}, this general incremental learning problem additionally faces an \textit{intra-class stability-plasticity dilemma} which refers to the trade-off between adapting novel examples and preserving current knowledge of the class, and an \textit{intra-class domain imbalance} problem, where the model is biased toward incoming domains due to a limited memory.
The intra-class problem is particularly challenging since the domain labels are usually unknown in practice.

The majority of current research on the class incremental learning either focuses on improving the inter-class stability-plasticity trade-off and imbalance issue~\cite{rebuffi2017icarl, delange2021continual} or mainly attempts to tackle the intra-class stability-plasticity dilemma~\cite{TaoHCG20, VolpiLR21, TangSCO21}. Recent works on the online general class incremental learning~\cite{LomonacoM17, BuzzegaBPAC20} typically ignore the intra-class structure of data distribution.
In particular, those methods usually adopt the same feature representation for the data from both incoming and existing domains of a class, which makes it difficult to learn new domains without interference with previously-learned representation of that class.
Such domain-invariant representations sacrifice the intra-class plasticity, often resulting in a poor intra-class trade-off between plasticity and stability.

In this work, we develop a novel domain-aware learning framework for the general incremental learning problem, which enables us to address both inter-class and intra-class challenges in a unified manner. To this end, we introduce a flexible class representation based on the von Mises-Fisher (vMF) mixture model to capture the intra-class structure and a bi-level balanced memory to cope with data imbalance within and across classes. 
In detail, we build a vMF mixture model on deep features of each class to learn a domain-aware representation and design an expansion-and-reduction strategy to dynamically increase the number of its components in new sessions. Combining with an inter- and intra-class forgetting resistence strategy like distillation, our design is capable of achieving better inter- and intra-class stability-plasticity trade-off. Moreover, based on the learned class representation, 
we propose a balanced memory at both inter- and intra-class level to mitigate bias toward new classes and new domains.

To learn our domain-aware representation, we devise an iterative training procedure for model update at each incremental session. Specifically, when new data comes, we first inherit the learned model from last session and allocate new components for the mixture model of each incoming category. 
We then adopt the Expectation-Maximization (EM) algorithm to jointly learn the backbone and mixture models, treating the component assignments of input data as latent variables. We incorporate strategies overcoming inter-class forgetting like \cite{hou2019learning, simonKH21, yan2021dynamically} and adopt intra-class knowledge distillation for alleviating inter- and intra-class catastrophic forgetting, respectively.
After each model update, we further perform a mixture reduction step based on hierarchical clustering to maintain a compact clustering result. 
During inference, we first extract input features via the backbone network and then infer its component assignment in class, followed by taking the class with the maximal component probability as prediction.

We validate our approach by extensive comparisons with prior incremental learning methods on three benchmarks: iDigits, iDomainNet and iCIFAR-20.
For each benchmark, we conduct experiments on splits with varying class and domain distributions over time.
The empirical results and ablation study show that our method consistently outperforms other approaches across all benchmarks.

In summary, the main contributions of our work are three-folds as follows:
\begin{itemize}
\item We formulate a new offline general incremental learning problem where both class distribution and intra-class domain distribution continuously change over time. This problem has the stability-plasticity dilemma and imbalance issue at both inter- and intra-class level.
\item We propose a method based on vMF mixture models to learn a domain-aware representation for addressing the general stability-plasticity dilemma and develop a bi-level balanced memory strategy to mitigate both the inter- and intra-class data imbalance issue.
\item Extensive experiments on three benchmarks show that our strategy consistently outperforms existing methods by a sizable margin.
\end{itemize}

\section{Related Works}\label{sec:relatedworks}
Existing literature in incremental learning can be summarized from three perspectives, including the problem settings, stability-plasticity dilemma and imbalance strategy.
\vspace{-7mm}
\paragraph{Problem Settings}
Most previous works focus on either class incremental learning~\cite{chaudhry2018riemannian, castro2018eeil, hou2019learning, yan2021framework, douillard2020podnet, tao2020tpcil, TaoHCG20, simon2021learning, rajasegaran2019random, yan2021dynamically} or domain incremental learning~\cite{TaoHCG20, VolpiLR21, TangSCO21}. Only a few attempts~\cite{aljundi2019online, BuzzegaBPAC20, LomonacoM17} address the general incremental learning problem, but these methods mainly learn from a data stream in an online fashion. 
In contrast, we study the offline learning setting, which allows multiple passes of incoming data at each incremental session.  
This paradigm is important in many real world applications~\cite{Pierre18,hu2022how} in which offline learning regularly outperforms the online version.
To the best of our knowledge, we are the first to tackle the general incremental learning that allows offline training in sessions.

Several recent studies tackle the class-incremental domain adaptation problem~\cite{kundu20cida, XuILR21cida}, which aims at adapting the model trained on a source domain to a target domain including novel classes. They mainly focus on the performance on the target domain and thus do not need to cope with the intra-class forgetting challenge. By contrast, our work requires the model to perform well on not only old and new classes, but also old and new domains of those classes.

It is also worth noting that while continual learning without data memory has attracted much attention in literature~\cite{yu2020semantic, zhu21nomem, verma21nomem}, these methods typically perform less effectively than those with a limited data memory for storing old examples~\cite{douillard2020podnet,yan2021dynamically}. 
In this work, we allow methods to access a finite number of previously seen examples as in the majority of existing incremental learning approaches.

\vspace{-3mm}
\paragraph{Stability-Plasticity Dilemma}
To alleviate the forgetting of learned representation, current continual learning methods can be largely grouped into three categories.
The first are the regularization-based methods \cite{kirkpatrick2017overcoming, chaudhry2018riemannian, ebrahimi2019uncertainty}, which add regularization on the parameters directly to prevent dramatic changes of the important parameters.  
The second are the distillation-based methods~\cite{castro2018eeil, hou2019learning, douillard2020podnet, tao2020tpcil, TaoHCG20, simon2021learning}, which adopt knowledge distillation to preserve the representation by penalizing the difference between outputs of previous and current models. 
The third are the structure-based methods~\cite{RajasegaranH0K019, yan2021dynamically}, which allocate new parameters at each new session and prevent the change of the representation learned at previous sessions.
However, the representations in these methods are usually domain invariant, which cannot provide enough intra-class plasticity without sacrificing the intra-class stability.
By contrast, our method can achieve better stability-plasticity dilemma by discovering and maintaining the intra-class structure.
\vspace{-3mm}
\paragraph{Imbalance Strategy}
The imbalance issue is largely caused by the limited size of memory.
To deal with this problem, most works~\cite{castro2018eeil, hou2019learning, wu2019bic, zhao2020wa} adjust the classifier weights or prediction logits between classes to eliminate the bias after learning the representations. 
We note that these works mainly focus on solving the inter-class imbalance problem and can not cope with or easily be extended to solve the intra-class domain imbalance due to the missing domain labels.
By contrast, our work can simultaneously achieve both inter- and intra-class balance with the help of domain label estimation in the EM framework.


\section{Method}\label{sec:method}
In this work, our goal is to address the general incremental learning problem in which both class and domain distribution change over time.
To this end, we propose to learn a domain-aware representation capable of achieving a better stability-plasticity trade-off at both intra- and inter-class level.
In particular, we develop a mixture model for each class to capture the intra-class structure and learn the mixture model via a new EM-based framework.

We first present the problem setup in ~\cref{subsec:setup}, followed by the presentation of model architecture in ~\cref{subsec:model_overview}. Then we introduce the adaptation of model at each session in ~\cref{subsec:train} and memory selection strategy in ~\cref{subsec:memory}, respectively. Finally, we describe the inference process in ~\cref{subsec:inference}.

\subsection{Problem Setup}{\label{subsec:setup}}
Firstly, we introduce the problem setup of general incremental learning. Formally, at session $t$, the model observes incoming data $\sD_{t} = \{(\vx_t^i, y_t^i)\}^{N_t}_{i=1}$ where $\vx_t^i \in \sX$ denotes the i-th image, $y_t^i\in\sC_{t}$ is its class label, and $N_t$ is the number of new examples.
Here we assume each class has multiple domains representing different variations in the class, e.g. background or style variation.
We denote the underlying domain label as $z_t^i$ for the data point $(\vx^i_t,y_t^i)$ and $z_t^i \in \sZ_t^c$ where $\sZ_t^c$ is its domain label space.
The class label space of the model is all observed classes $\tilde{\sC}_{t} = \cup_{i=1:t} \sC_{i}$, and the domain label space is $\tilde{\sZ}^{\ry}_{t} = \cup_{i=1:t} \sZ^{\ry}_{i}$ for class $\ry$.
It is worthy noting that $P({\sC}_t \cap \tilde{\sC}_{t-1} \neq \emptyset) > 0$ and $P(\sZ^{\ry}_t \cap \tilde{\sZ}^{\ry}_{t-1} \neq \emptyset) > 0$, which means previously observed categories or domains may repeatedly appear in the subsequent sessions. 
Given a loss function $\mathcal{L}(y,\hat y)$ where ${y}$ is ground truth and $\hat y$ is the label prediction, the risk associated with the model $\mathcal{M}$ is defined as follows
\begin{equation}\label{eq:idealtarget}
	\mathbb{E}_{\ry \sim P(\ry|t)}[ \mathbb{E}_{\rz \sim P(\rz|\ry, t)} [ \mathbb{E}_{\rx \sim p(\rx|\ry,\rz, t)}[\mathcal{L}(y,\hat y)]]]
\end{equation}
where $P(\ry|t)$ denotes the class distribution on $\ry \in \tilde{\sC}_t$, $P(\rz|\ry, t)$ refers to the class-specific domain distribution on $\rz \in \tilde{\sZ}_t^{\ry}$, and $p(\rx|\ry,\rz,t)$ is the conditional data generation distribution given class $\ry$ and domain $\rz$. 
For simplicity, we assume $p(\rx|\ry,\rz,t)$ does not change over sessions in this work, which often holds in real-world scenarios. 
At session $t$, due to memory limitation, model can only keep a small subset of the dataset, which is denoted as memory $\sM_{t+1}$. 
The data available for training at session $t$ is the union of $\sD_{t}$ and $\sM_{t}$, denoted as $\tilde{\sD}_{t}=\sD_{t} \cup \sM_{t}$. 
For notation clarity, we omit the subscript $t$ in the following subsections.

\subsection{Model Architecture}{\label{subsec:model_overview}}
At session $t$, our model $\mathcal{M}$ consists of a backbone network $\mathcal{F}$ with parameters $\vtheta$ and a mixture model with parameters $\bm{\phi}$. 
We denote the parameters of our entire model as $\Theta=\{\vtheta, \bm{\phi}\}$.
Concretely, given an image $\vx$, we extract the feature $\vv = \mathcal{F}_{\vtheta}(\vx)$.
We perform $\normltwo$ normalization on the feature $\vv$ and obtain the unit length feature vector $\tilde{\vv} = \vv/{\lVert \vv \rVert}$ , in order to alleviate the imbalance issue (following the practices in \cite{hou2019learning}).
For each class, we model the feature distribution over $\tilde{\rvv}$ with a mixture model as follows: 
\vspace{-5mm}
\begin{align}
    \vspace{-4mm}
	p(\tilde{\rvv}|\ry) = \sum_{k=1}^{K_\ry} P(\rz=k|\ry) p(\tilde{\rvv} | \rz=k, \ry)
\end{align}
where $K_\ry$ is the number of components in the mixture model of class $\ry$, and $P(\rz|\ry)$ represents the component proportions, which follows a multinomial distribution.
In practice, we set the distribution $P(\rz=k|\ry) = 1/K_\ry$ as uniform distribution to mitigate the intra-class domain imbalance issue.
Moreover, the probability density function $p(\tilde{\rvv} |\rz, \ry) = C_{d}(\kappa) e^{\kappa \tilde{\vmu}_{\ry,\rz}^{\top} \tilde{\rvv}}$ follows the von Mises-Fisher(vMF) distribution~\cite{BanerjeeDGS05}, which can be considered as multivariate normal distribution for directional features on the hyper-sphere.
Concretely, The concentration parameter $\kappa \geq 0$, $d \geq 2$, and the normalization coefficient $C_{d}(\kappa) = \frac{\kappa^{d/2-1}}{(2\pi)^{d/2}I_{d/2-1}(\kappa)}$ with $I_{r}(\cdot)$ represents the modified Bessel function of the first kind and order $r$.
Note that we assume every component shares the same $\kappa$ for convenience in this work.



\begin{figure*}[t]
	\includegraphics[width=\textwidth]{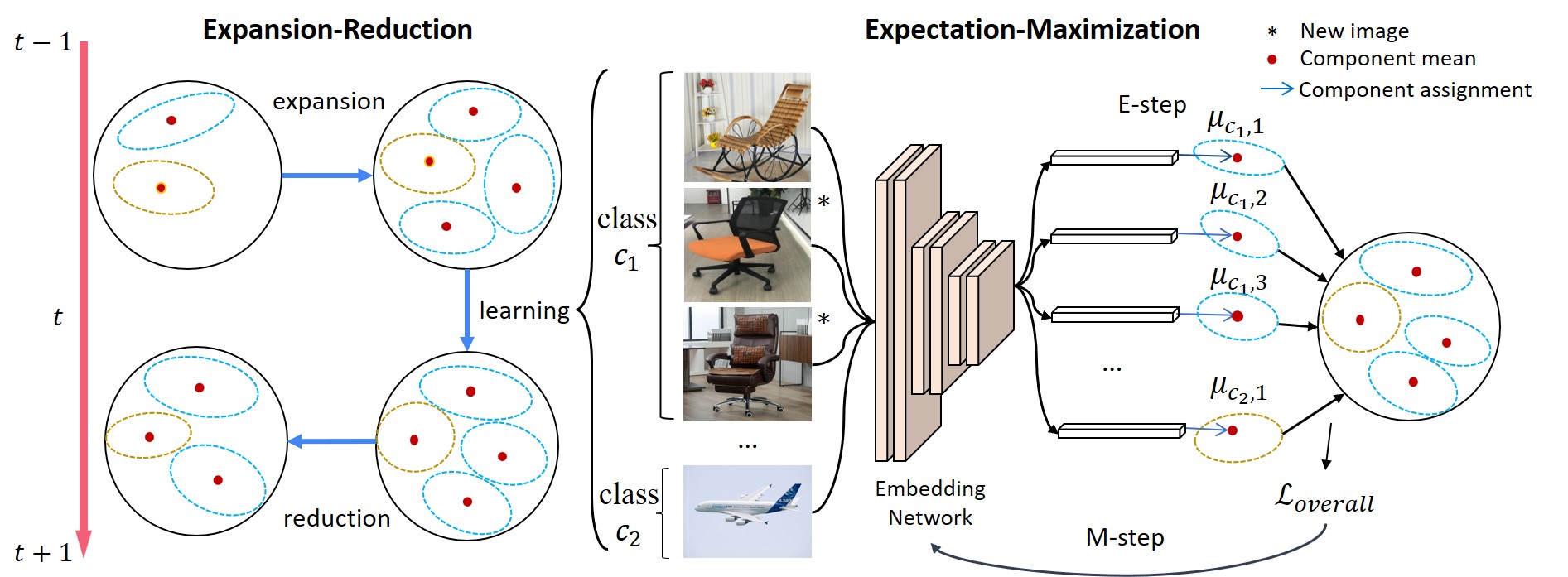}
	\caption{\textbf{Method Overview:} At session $t$, the model starts with the state at last session. It observes incoming images and map them into unit hyper-sphere in the feature space. For the classes in the incoming categories like $c_1$ in the figure, we first expand the mixture model for each incoming class, followed by the learning with an EM framework. In E-step, we perform a component assignment by choosing the component with the closest mean $\vmu$. In M-step, we update both the embedding network and mixture models with overall loss. After the learning, we perform a mixture model reduction to reduce the redundant components for each class.}
	\label{fig:method_overview}
\end{figure*}

\subsection{Model Adaptation}{\label{subsec:train}}
We now introduce our incremental learning strategy (see Fig.~\ref{fig:method_overview} for an overview). Specifically, to update the model in a session, we first develop an expansion-and-reduction strategy to dynamically determine the number of components in the mixture model. This enables the model to better accommodate new distributions especially when the number of domains in a class changes.
Given a model structure, we then introduce an EM-based framework to learn the mixture model by treating its component assignments as latent variables.
Concretely, we expand the mixture models at the beginning of each session, and then jointly learn the backbone and mixture models using the EM framework. Finally, we perform mixture model reduction to maintain a compact representation.

\vspace{-2mm}
\paragraph{Expansion-and-Reduction}
As the number of new domains for each class is unknown, we first increase the number of components and then use a reduction step to obtain the final number of components after the model training.
At the beginning of each session $t$, the model $\mathcal{M}$ is inherited from last session and then expanded with new components and/or mixture models.
Concretely, for each class $y \in \sC_{t}$, we add $m$ components to 
the corresponding mixture model if $y$ has been encountered previously (i.e., $y\in\mathbb{C}_{t-1}$), or create a new mixture model with $m$ components if $y$ is a new class, where the newly-added components are randomly initialized and $m$ is a large number. 
After expansion, the feature distribution over $\tilde{v}$ becomes
\begin{align}
	p(\tilde{\rvv}|\ry) = \sum_{k=1}^{K_{\ry}^{t-1} + m} P(\rz=k|\ry) p(\tilde{\rvv} | \rz=k, \ry)
\end{align}
where $K_{\ry}^{t - 1}$ is the final number of components of class $y$ at session $t-1$ and $P(\rz=k|\ry)$ remains a uniform distribution.

After the model learning (as described below), we perform a mixture model reduction step to avoid over-segmenting the feature space with redundant components or increasing the overall model complexity. 
Specifically, we group the vMF components and re-represent each group by a new single vMF density.
We can view vMF component clustering as a standard data clustering with additional requirement that data points sharing the same original vMF component should end up in the same output component.  
In practice, we adopt a hierarchical clustering on the original components, which works reasonably well empirically.
We treat every component as a distinct cluster and then recursively merge pairs of clusters with distance smaller than a predefined threshold $\delta$ within each class. 
To merge the paired clusters $i$ and $j$, the mean $\tilde{\vmu}$ of the new component's vMF density is updated as follows
\begin{align}
    \vmu = \frac{1}{n_i+n_j} (\sum_{l=1}^{n_i} \tilde{v}^i_l + \sum_{q=1}^{n_j} \tilde{v}^j_q),\quad 
    \tilde{\vmu} = \frac{\vmu}{||\vmu||} 
\end{align}
where $n_i$, $n_j$ are the number of examples in the cluster $i$ and $j$, respectively. $\tilde{v}^i_l$ is the normalized feature of $l$-th example in the $i$-th cluster.
\vspace{-2mm}
\paragraph{Model Learning}
To learn the domain-aware categorical representation, we develop an EM learning algorithm to train the entire model $\mathcal{M}$ with a limited memory, in which the component assignments $z$ of image $\vx$ are treated as latent variables.
The log-likelihood of a given data $\vx$ can be written as
\begin{gather}\label{eq:elbo}
	\begin{split}
	\log P(\ry|\vx,\Theta)
		\ge \E_{Q(\rz)}\left[\log \frac{P(\rz,\ry|\vx;\Theta)}{Q(\rz)}\right]	
	\end{split}
\end{gather}
where the right hand $\E_{Q(\rz)}\left[\log \left(P(\rz,\ry|\vx;\Theta)/Q(\rz) \right) \right]$ is the evidence lower bound (ELBO).


\noindent\textbf{E-step} 
In the E-step, we compute a new estimate of the component assignments using the learned parameters $\Theta'$ from the last M-step, represented as $Q^*(\rz)$. 
Concretely, $Q^*(\rz) = \llbracket\rz=\hat z\rrbracket$ where $\llbracket\cdot\rrbracket$ is the indicator function and $\hat{z}$ is defined as  
\begin{gather}\label{eq:assign}
	\begin{split}
	\hat z &= \mathop{\arg\max}_{k} P(\rz = k|\rvx, \ry; \Theta' ) \\
		   &= \mathop{\arg\max}_{k} \frac{p(\vx|\rz=k, \ry;\Theta') P(\rz=k|\ry)}{\sum_{l=1}^{K_{\ry}}p(\vx|\rz=l, \ry; \Theta')P(\rz=l|\ry)} \\
		   &= \mathop{\arg\max}_{k} \tilde{\vmu}_{\ry,k}^{\top} \tilde{\vv}
	\end{split}
\end{gather}
where $\tilde{\vmu}_{\ry,k}$ is the mean of the k-th component for class $\ry$.
In other words, we update the component assignment of image $\vx$ within  class $\ry$ by taking the component with the closest mean feature $\tilde{\vmu}$ on the hyper-sphere. In contrast to the standard E-step using the posterior $P(\rz|\ry,\vx;\Theta')$ as $Q(\rz)$, our method can be viewed as a hard-EM approximation.

%

\noindent\textbf{M-step} 
In the M-step, we maximize the ELBO by mini-batch SGD based on the component assignments obtained in the E-step. 
This optimization problem can be rewritten as follows
\begin{align}
	\min_{\Theta} \E_{(\rx,\ry)\sim p(\rx,\ry)}[&\mathop{KL}\left[Q^*(\rz) || P(\rz|\ry,\vx;\Theta) \right]  \\ 
	&- \log P(\ry|\vx;\Theta)],\nonumber
\end{align}
which enforces the model to learn the classification at both inter- and intra-class level.
Moreover, given dataset $\tilde{\sD}$, this expectation can be rewritten as follows
\begin{align}\label{eq:loss}
	\mathcal{L}_{\text{clf}} &= \mathcal{L}_{\text{clf}}^{\text{inter}} + \lambda \mathcal{L}_{\text{clf}}^{\text{intra}}  \\
	&= -\frac{1}{|\tilde{\sD}|}\sum_{i=1}^{|\tilde{\sD}|} \big(\log P(\ry=y_{i}|\vx_i;\Theta) \nonumber\\
	&+ \lambda \log P(\rz=\hat z_{i}|\vx_i,y_{i};\Theta) \big)
\end{align}
where $\mathcal{L}^{\text{inter}}_{\text{clf}}$ is the inter-class classification loss, $\mathcal{L}^{\text{intra}}_{\text{clf}}$ refers to the intra-class classification loss, $\lambda$ is the hyper-parameter to balance these two losses, and the posterior of component assignment is computed as follows
\begin{align}
P(\rz=k|\ry,\vx;\Theta) = \frac{e^{\kappa \tilde{\vmu}_{\ry,k}^{\top} \tilde{\vv}}}{\sum_{l=1}^{K_{\ry}}e^{\kappa \tilde{\vmu}_{\ry,l}^{\top} \tilde{\vv}}}.
\end{align}
It is worth noting that the component assignment $\rz$ depends on the prediction of label $\ry$.
Consequently, we design a schedule of $\lambda$ which starts with zero and gradually grows over iterations as the quality of label prediction $\ry$ improves.
Moreover, to maintain inter-class class balance, we assume class distribution $P(\ry)$ follows uniform distribution, and then the prediction probability is given by 
\begin{gather}\label{eq:pred}
	\begin{split}
		P(\ry=c|\vx;\Theta) &= \frac{\sum_{i=1}^{K_c} p(\vx,\rz=i|\ry=c;\Theta)P(\ry=c)}{\sum_{m=1}^{|\tilde{\sC}_{t}|} \sum_{n=1}^{K_{m}} p(\vx,\rz=n,\ry=m;\Theta)} \\
		&= \frac{\frac{1}{K_c}\sum_{i=1}^{K_c}e^{\kappa \tilde{\vmu}^{\top}_{c,i} \tilde{\vv}}}{\sum_{m=1}^{|\tilde{\sC}_{t}|}\sum_{n=1}^{K_{m}} \frac{1}{K_m}e^{\kappa \tilde{\vmu}_{m,n}^{\top} \tilde{\vv}}}
	\end{split}
\end{gather}

To prevent intra-class forgetting and preserve the learned intra-class structure, we employ a knowledge distillation loss within each class, which is defined as 
\begin{gather}\label{eq:distill_loss}
\begin{split}
	\mathcal{L}_{\text{dis}} = \frac{1}{|\tilde{\sD}|}\sum_{i=1}^{|\tilde{\sD}|} \frac{1}{|\tilde{\sC}|}\sum_{c=1}^{|\tilde{\sC}|}\mathop{KL}(&P(\rz|\ry=c, \vx_i;\Theta)||\\
	&P(\rz|\ry=c, \vx_i;\Theta_{\text
	{old}}))
\end{split}
\end{gather}
where $\Theta_{\text{old}}$ is the parameters of learned model from previous session.
Furthermore, we introduce a component regularization loss \cite{shah2017robust} on the mixture model of each class to learn a tight cluster, which is computed as below
\begin{align}
	\mathcal{L}_{\text{reg}} = -\frac{1}{|\tilde{\sC}|} \sum_{y \in \tilde{\sC}} \sum_{i=1}^{K_y} \sum_{j=i+1}^{K_y} \frac{1}{K_y*(K_y-1)} \tilde{\vmu}_{y,i}^{\top}\tilde{\vmu}_{y,j}
\end{align}
Since we uses many components at the beginning of each session to expand current model, this loss can prevent the model from over-segmenting the feature space.

Finally, the overall loss function in the M-step is a linear combination of those three losses, defined as follows 
\begin{align}\label{eq:final_loss}
\mathcal{L}_{\text{overall}} = \mathcal{L}_{\text{clf}} + \beta \mathcal{L}_{\text{dis}} + \eta \mathcal{L}_{\text{reg}}
\end{align}
where $\beta, \eta$ are the loss weighting coefficients.



\subsection{Memory Selection}{\label{subsec:memory}}
We introduce a bi-level balanced strategy to build the data memory $\mathbb{M}_t$, which maintains a class-balanced and domain-balanced replay dataset in each session.
Concretely, we first assign $m=\mathcal{B}/|\tilde{\sC}|$ exemplars for each class to ensure the inter-class class balance, where $\mathcal{B}$ is the maximum number of exemplars that can be saved. 
Subsequently, given the mixture model for each class $c$, we select $m/|K_c|$ samples from each component of class $c$ uniformly to achieve an intra-class domain balance.
We note that such a strategy aims to achieve better average performance. 

\subsection{Model Inference}{\label{subsec:inference}}
Given a new image $\mathbf{x}$, the model inference computes its normalized feature $\tilde{\vv}$ and predicts the label $\hat{y}$ by taking the class of the closest component in the feature space, which can be written as
\begin{equation}
	\hat{y} = \mathop{\arg\max}_{c} \max_{k} \ \tilde{\vmu}_{c,k}^{\top} \tilde{\vv} 
\end{equation}


\section{Experiments}\label{sec:exps}
We conduct a series of experiments to verify the effectiveness of our method. 
In this section,  we first introduce the experiment setup including the benchmarks, types of distribution shifts and the comparison methods in~\cref{subsec:benchmark}, followed by the implementation details in~\cref{subsec:detail}.
Then we show our experimental results in~\cref{subsec:result}. 
In the end, we demonstrate the analysis of our method to provide more insights in~\cref{subsec:analysis}.
\begin{table*}[t]
	\caption{\textbf{Results:} Average incremental accuracy(\%) over sessions on iCIFAR-20, iDomainNet, and iDigits with three representative splits. 's' represents the number of sessions in the split. E.g. 5s means this split has 5 sessions.}
	\centering
	\resizebox{0.95\textwidth}{!}{
	\begin{tabular}{l|lll|lll|lll}
		\toprule
		\multirow{2}{*}{Methods} & \multicolumn{3}{c|}{iCIFAR-20}                                                & \multicolumn{3}{c|}{iDomainNet}                                               & \multicolumn{3}{c}{iDigits}                                                 \\ \cmidrule{2-10} 
		& \multicolumn{1}{c}{NC (5s)} & \multicolumn{1}{c}{ND (5s)} & \multicolumn{1}{c|}{NCD (10s)} & \multicolumn{1}{c}{NC (10s)} & \multicolumn{1}{c}{ND (6s)} & \multicolumn{1}{c|}{NCD (10s)} & \multicolumn{1}{c}{NC (5s)} & \multicolumn{1}{c}{ND (4s)} & \multicolumn{1}{c}{NCD (10s)} \\ \midrule
		Replay                            & 74.22 & 72.47 & 67.56 & 45.73 & 47.40 & 42.74 & 83.22 & 92.75 & 80.24 \\
		iCaRL~\cite{VolpiLR21}            & 78.98 & 73.06 & 70.90 & 51.64 & 48.40 & 44.60 & 89.03 & 93.26 & 85.12 \\
		EE2L~\cite{hou2019learning}       & 78.50 & 73.86 & 70.52 & 52.03 & 48.03 & 43.54 & 89.97 & 93.91 & 85.46 \\
		Meta-DR~\cite{VolpiLR21}          & 73.22 & 71.09 & 66.65 & 46.40 & 48.73 & 44.15 & 89.89 & 94.00 & 86.31 \\ \midrule
	    UCIR~\cite{hou2019learning}       & 78.19 & 76.01 & 72.54 & 50.17 & 49.25 & 44.53 & 89.41 & 94.17 & 86.29 \\
		UCIR w/ ours   & 78.82 & 78.08 & 75.62 & 50.52 & 52.85 & 49.81 & 90.51 & 95.50 & \textbf{90.42} \\ \midrule
		GeoDL~\cite{simonKH21}            & 78.92 & 76.43 & 72.96 & 51.64 & 49.33 & 45.12 & 89.72 & 93.98 & 86.24 \\ 
		GeoDL w/ ours                     & 79.49 & 79.40 & 76.19 & 51.81 & 53.32 & 51.20 & 89.86 & 94.80 & 89.82  \\ \midrule
		DER~\cite{yan2021dynamically}     & 82.17 & 74.87 & 74.56 & 66.58 & 46.76 & 50.00 & 89.01 & 93.62 & 84.80  \\
		DER w/ ours                       & \textbf{82.52} & \textbf{84.03} & \textbf{82.11} & \textbf{66.85} & \textbf{61.05} & \textbf{57.07} & \textbf{91.32} & \textbf{97.07} & 88.65 \\ \bottomrule
	\end{tabular}}
	\label{tab:results}
\end{table*}

\begin{figure*}[t]
	\centering
	\includegraphics[width=0.98\textwidth]{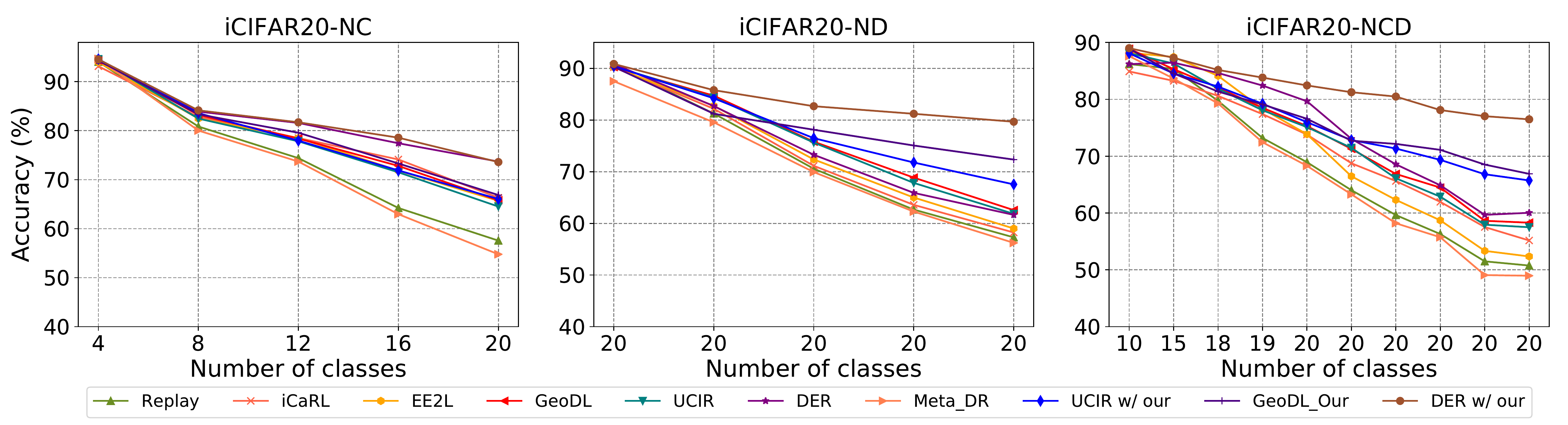}
	\caption{\textbf{Performances w.r.t sessions} on iCIFAR-20 benchmark with three splits}
	\label{fig:steps_perf}
	\vspace{-4mm}
\end{figure*} 

\subsection{Experiment Setup}{\label{subsec:benchmark}}

We conduct experiments on three benchmarks, including iDigits, iDomainNet and iCIFAR-20:
\begin{itemize}
    \item \textit{iDigits:} We follow \cite{VolpiLR21} to construct a digit recognition benchmark, which includes four datasets: MNIST \cite{lecun98mnist}, SVHN \cite{Netzer11svhn}, MNIST-M \cite{GaninL15} and SYN \cite{GaninL15}. Each dataset represents a distinct domain. 
	\item \textit{iDomainNet:} It is constructed from DomainNet \cite{PengBXHSW19}, a well-known dataset for domain adaptation. It contains six domains, which are \textit{Clipart}, \textit{Infograph}, \textit{Painting}, \textit{Quickdraw}, \textit{Real} and \textit{Sketch}. Each domain contains 345 categories of common objects.
	Because some domains in these classes contains only few images($\sim10$), we select the top 100 classes with most images, which contains 132,673 training data in total.
	The smallest domain in these 100 classes has 52 images.
	\item \textit{iCIFAR-20:} It is based on CIFAR-100 \cite{krizhevsky2009learning}, which has 20 super classes and 5 subclasses for each super class. These subclasses are considered to be different domains of the same class and model needs to predict super class labels in the recognition task.
\end{itemize}
Each domain of the iCIFAR-20 has the same number of training images. By contrast, for iDomainNet and iDigits, each domain has a different number of training images.
We evaluate these methods using three representative splits to simulate different scenarios for every benchmark, in which the distributions of classes and domains shift:
\begin{itemize}
	\item \textit{New Class(NC)}: Incoming data contains images from new categories only.
	\item \textit{New Domain(ND)}: Incoming data contains images from new domains only. 
	\item \textit{New Class and Domain(NCD)}: Incoming data contains images from new categories or new domains.
\end{itemize}
\begin{figure*}[t]
	\centering
	\begin{subfigure}{0.49\textwidth}
		\includegraphics[width=\textwidth]{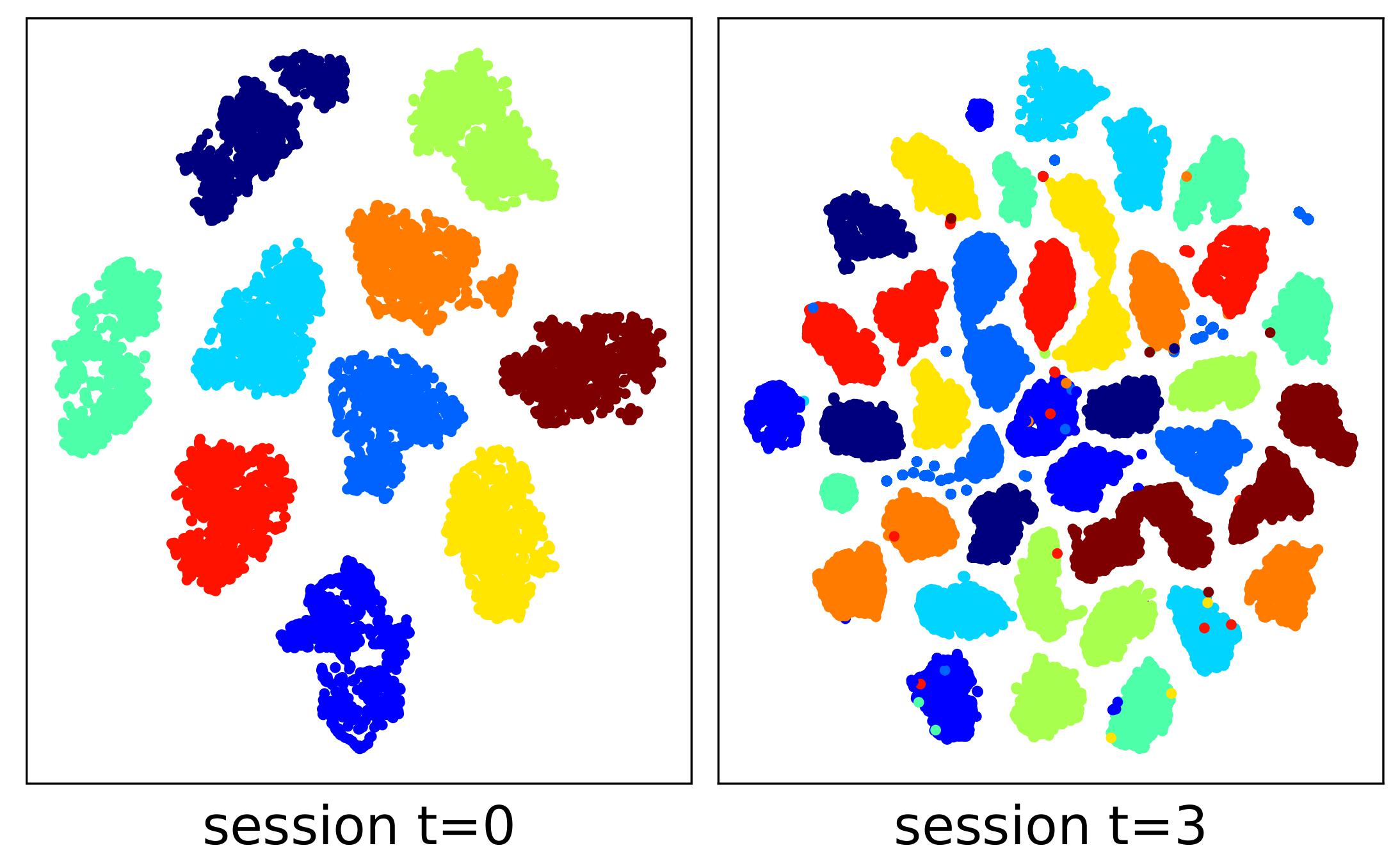}
		\caption{t-SNE on images across classes}
	\end{subfigure}
	\begin{subfigure}{0.49\textwidth}
		\includegraphics[width=\textwidth]{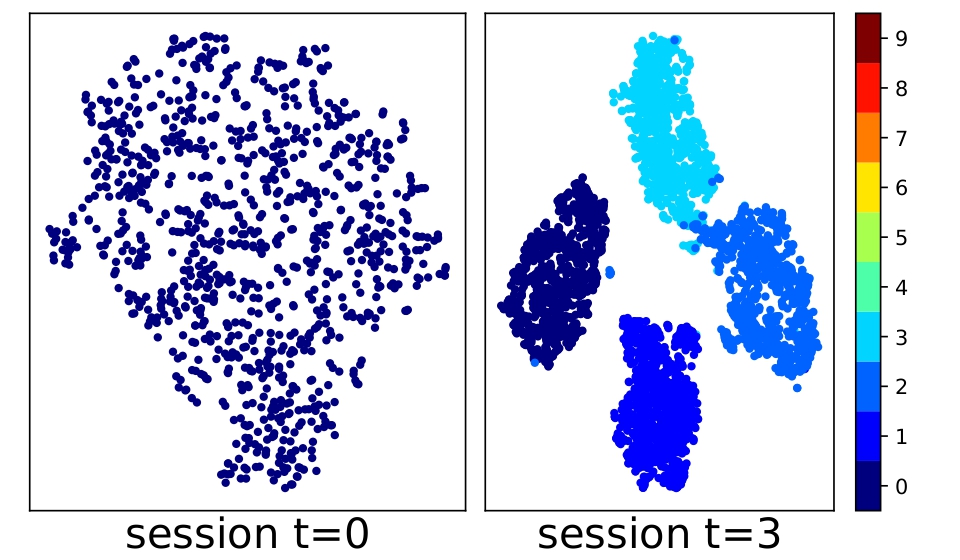}
		\caption{t-SNE on images of digit 0}
	\end{subfigure}
	\caption{\textbf{t-SNE Visualization} on all data so far seen of DER w/ ours across sessions for iDigits NC split. Different colors represent class label for the left and domain label for the right.}
	\label{fig:tsne}
\end{figure*}
 \begin{table*}[t]
	\caption{\textbf{Ablation Study:} Contribution of each component evaluated on iCIFAR-20.}
	\centering
	\resizebox{\textwidth}{!}{
	\begin{tabular}{cccc|cccccc}
		\toprule
		\multicolumn{4}{c|}{\textbf{Components}}  & \multicolumn{2}{c}{\textbf{NC}} & \multicolumn{2}{c}{\textbf{ND}} & \multicolumn{2}{c}{\textbf{NCD}} \\
		\textbf{Mixture model} & \textbf{Expansion-Reduction} & \textbf{component regularization Loss} & \textbf{Bi-level Memory} & \textbf{Final($\%$)} & \textbf{Avg($\%$)}   & \textbf{Final($\%$)} & \textbf{Avg($\%$)}   & \textbf{Final($\%$)} & \textbf{Avg($\%$)}    \\
		\midrule
		\xmark & \xmark    & \xmark        & \xmark              &   72.06        &      82.07     &     67.35      &    77.18       &    54.32       &   70.64         \\
		\checkmark  & \xmark  & \xmark      & \xmark              &   72.39       &    81.38  &     68.98      &        80.39     &       68.47     &     78.71      \\
		\checkmark  & \checkmark  & \xmark      & \xmark  &  73.95 &  82.32  &  70.32   &  81.28 &    69.86  &   80.26           \\
		\checkmark  & \checkmark  & \checkmark      & \xmark              &   74.14      &   82.46  &    71.92      &     82.04           &   71.56        &     80.97         \\
		\checkmark  & \checkmark  & \checkmark      & \checkmark          &   \textbf{74.49}  &   \textbf{82.77}     &     \textbf{80.13}      &   \textbf{84.11} &    \textbf{73.70}        &   \textbf{82.17}   \\
		\bottomrule
	\end{tabular}}
	\vspace{-4mm}
	\label{tab:ablation}
\end{table*}


For the NC split, we construct iCIFAR20-NC and iDigits-NC by splitting iCIFAR-20 and iDigits into 5 sessions with 4 and 2 classes per session, respectively.
Moreover, the model is trained in batches of 60 classes with 10 sessions in total on iDomainNet-NC.
For the ND split, every session has all the classes and each class has one incoming domain, where iCIFAR-20, iDigits and iDomainNet are split into 5, 4 and 6 sessions, respectively.
For the NCD split, we divide all domains in the dataset into ten sessions for each of the three datasets. For more information about these splits, please refer to the appendix.

We adopt the Replay, iCaRL~\cite{rebuffi2017icarl}, EE2L~\cite{castro2018eeil}, UCIR~\cite{hou2019learning}, 
GeoDL~\cite{simonKH21}, DER~\cite{yan2021dynamically} and Meta-DR~\cite{VolpiLR21} as the comparison methods.
Here the Replay refers that the model is fine-tuned using both the memory and incoming data.
It is noteworthy that iCaRL, EE2L, UCIR, GeoDL and DER are all designed to address the class incremental learning problem, whereas Meta-DR are proposed to resolve domain incremental learning problem.
In contrast, our method mainly learns the intra-class structure and can be utilized by any class incremental learning method to address the stability-plasticity dilemma at the inter- and intra-class level. 
Consequently, we combine our method with three existing incremental learning approaches, UCIR, GeoDL and DER, in our experimental evaluation.


\vspace{-3mm}
\subsection{Implementation Details}{\label{subsec:detail}}
All these methods are implemented with PyTorch~\cite{paszke2017automatic}. 
We resize the images in iCIFAR-20 and iDigits to 32x32 and the images in iDomainNet to 112x112.
For the iCIFAR-20 and iDomainNet benchmarks, we follow DER~\cite{yan2021dynamically} and adopt the standard ResNet18~\cite{he2016deep} architecture as the feature extractor. We use SGD optimizer to train the network with 200 epochs in total for each session. Learning rate starts with 0.1 and is reduced by 0.1 at 80 and 120 epoch. 
We set the fixed memory size for these two benchmarks as 2000 instances. 
For the iDigits benchmarks, we choose the modified 32-layer ResNet used in \cite{rebuffi2017icarl, hou2019learning} because it is a simple dataset and large networks can be easily overfitted.
We train the methods with SGD optimizer for 70 epochs for each session, beginning with learning rate 0.1, which is reduced by 0.1 at 48 and 63 epochs. 
We set the fixed memory size as 500 for iDigits. 
In addition, the batch size is selected as 128 for iCIFAR-20 and iDigits, and 256 for iDomainNet. 
Weight decay is 0.0005 for all benchmarks. The distillation loss coefficient $\beta=1$.
The coefficient $\lambda$ in~\cref{eq:loss} linearly increases from 0 to 0.1 for the first 10 epochs and then is fixed at 0.1. The coefficient $\eta$ of the regularization loss is set as 0.1. 
For the expansion of mixture model, the number of components $m$ to add for each class is set as 30 for all benchmarks.
For the reduction of mixture model, the threshold $\delta$ is chosen as 0.7 for all benchmarks.
We run the E-step to update the component assignments at the beginning of each epoch.
Following \cite{douillard2020podnet}, these hyper-parameters are tuned on a val set built from the original training data.

\begin{table*}[t]
	\caption{\textbf{Sensitive Study} Influence of threshold $\delta$ in mixture model reduction on our method for iCIFAR-20 NCD split.}
	\centering
	\resizebox{\textwidth}{!}{
		\begin{tabular}{c|cc|cc|cc|cc|cc|cc|cc}
			\toprule
			\multicolumn{1}{c|}{\multirow{2}{*}{Threshold}} & \multicolumn{2}{c|}{$\delta=0.5$} & \multicolumn{2}{c|}{$\delta=0.55$} & \multicolumn{2}{c|}{$\delta=0.6$} & \multicolumn{2}{c|}{$\delta=0.65$} & \multicolumn{2}{c|}{$\delta=0.7$} & \multicolumn{2}{c}{$\delta=0.75$} & \multicolumn{2}{c}{$\delta=0.8$} \\
			\multicolumn{1}{c|}{}                             & Final         & Avg          & Final         & Avg          & Final         & Avg          & Final         & Avg          & Final         & Avg          & Final         & Avg   & Final & Avg       \\ \midrule
			DER w/ ours                                       & 74.08         & 80.70        & 75.24         & 81.19       & 75.47  & 81.21 & 75.74      & 81.26        & 76.5         & 82.17        & 77.44         & 82.05        & 76.49         & 82.01     \\   
			\bottomrule
	\end{tabular}}
	\label{tab:threshold}
\end{table*}
\subsection{Experimental Results}{\label{subsec:result}}
~\cref{tab:results} summarizes the average accuracy over sessions of different methods.
We combine our methods with three different methods - UCIR, GeoDL and DER, since they performs better than other baselines on at least one split. 

On all kinds of data distribution shifts in each benchmark, our method consistently improve the performance of these three methods, which demonstrates its effectiveness.
Particularly, we can see that our method solves the performance bottleneck of DER on the ND split and DER w/ ours consistently achieves the highest average accuracy in the majority of cases, e.g. $84.11\%$ on iCIFAR20-ND. Besides, UCIR w/ ours performs best on iDigits-NCD split with $90.42\%$ accuracy.
As demonstrated in ~\cref{fig:steps_perf}, we observe that our method consistently performs better than other method at each session for different splits. 
Specifically, the final session accuracy is boosted from $60.04\%$ to $76.40\%(+\bm{16.36}\%)$ on iCIFAR20-NCD split by incorporating our method into DER. 
Additionally, we find that integrating our method can significantly increases the performance of existing class incremental learning methods such as UCIR and DER on ND and NCD splits, while maintaining comparable performance on the NC split.

With regard to the ND split, UCIR and GeoDL perform better than other baselines, which is because their distillation is based on features. However, iCaRL and EE2L, which use logit-based distillation, perform worse because they penalize the change of predictions for data of old classes from new domains. As for DER, the old feature extractors cannot recognize data of old classes in new domains, which affects its prediction. For the NCD split, DER is the best for iCFIAR-20 and iDomainNet, since it performs much better on the NC split. Furthermore, Meta-DR performs well on the iDomainNet and iDigits but does not perform well on iCIFAR-20. This is because one domain in iCIFAR-20 represents one semantic subclass and domain randomization cannot solve the this domain gap.  



\subsection{Analysis}{\label{subsec:analysis}}

\paragraph{Ablation Study}
~\cref{tab:ablation} summarizes the results of our ablative experiments on iCIFAR-20, starting with DER. We can find that our method achieve $8.07\%$ improvement on average incremental accuracy for the NCD split by mixture model. We also show that the performance of the model is consistently improved over three different types of distribution shifts with our expansion and reduction strategy, especially achieving $0.89\%$ gain on the ND split. Furthermore, it shows that we can obtain $0.76\%$ improvement after adding component regularization loss. Finally, our method further improve the accuracy by $2.07\%$ for the ND split and $1.20\%$ for the NCD split, after adding the bi-level memory sampling approach.


\vspace{-2mm}
\paragraph{Visualization}
We utilize t-SNE~\cite{mattten08tsne} to visualize the feature embeddings on the iDigits ND split at different sessions, shownd in~\cref{fig:tsne}. 
As the number of sessions increases, each cluster mainly contains only examples from  the domain, implying a high degree of purity for each cluster.
It is noteworthy that each class in this split has four domains in the last session ($t=3$) and our method can separate most classes into four groups.
Furthermore, we take images of one class for further analysis, shown on the right side of ~\cref{fig:tsne}. 
It reveals our method is able to assign most instances to their respective domain labels, demonstrating the effectiveness of latent variable estimation.
\vspace{-2mm}
\paragraph{Sensitive Study}
We conduct sensitive study on the influence of threshold $\delta$ in the mixture model reduction step, as shown in~\cref{tab:threshold}, which shows our method is robust to small variations of the threshold. We also study the influence of different memory sizes and number of newly added components $m$, which are shown in the appendix.


\section{Conclusion and Discussion}~\label{sec:conclusion}
In this work, we propose and formulate the offline general incremental learning problem, which has many real-world applications. To address the challenges, we introduce a domain-aware learning framework.
Concretely, we propose a flexible class representation based on the mixture model to solve the stability-plasticity dilemma, which is learned by an expansion-reduction strategy and the EM algorithm. Furthermore, we also develop a bi-level balanced memory selection strategy based on the learned mixture model for the imbalance challenge.
We conduct exhaustive experiments on three benchmarks to validate the effectiveness of our method.
The experimental results demonstrate that our method consistently outperforms than other methods on three representative splits for each benchmark.
Furthermore, it is meaningful to apply our method to other vision tasks like video classification~\cite{LomonacoM17} and semantic segmentation~\cite{DouillardCDC21} as future work.

\paragraph{Limitations and Negative Impact}
Our method is designed for allowing access to limited old examples, and cannot be used to the setting without memory. As our method continuously updates the model with incoming data, it can be leveraged by malicious applications to upgrade its model with new data.


{\small
    \bibliographystyle{ieee_fullname}
    \bibliography{egbib}
}

\clearpage

\begin{appendices}
	\section{More Implementation Details}
We use Nvidia Titan XP as the computation platforms with CUDA 10.1. Our python is 3.7 and PyTorch is 1.71.
We use seed 1993 for all the experiments.

\paragraph{Data Augmentation} 
For iDigits benchmark, we only resize every image to 32x32. For iCIFAR-20 benchmark, we use RandomCrop with shape 32x32 and padding 4. We also use RandomHorizontalFlip, ColorJitter with brightness as 63/255 and Normalization with mean (0.5071,0.4867,0.4408) and standard deviation (0.2675,0.2565,0.2761). For iDomainNet, We apply RandomResizedCrop with size as 112 , RandomHorizontalFlip and Normalization with mean (0.5,0.5,0.5) and standard deviation (0.5,0.5,0.5) for all the channels.

\section{Details of Splits}
We introduce the NCD splits for the three benchmarks.
For iDigits NCD split, numbers of new classes at each session are [4, 3, 2, 1, 0, 0, 0, 0, 0, 0].
For iDomainNet NCD split, numbers of new classes at each session are [60, 10, 10, 10, 10, 0, 0, 0, 0, 0].
For iCIFAR-20 NCD split, numbers of new classes at each session are [10, 5, 3, 1, 1, 0, 0, 0, 0, 0].
The number of new classes decreases as the number of session increases because unseen categories for models in real world should become fewer and fewer with the accumulation of knowledge.
Each class at every session contains data from one domain. We will release the data of the three benchmarks with all the splits used in the experiments later.

\section{More Curves}

\begin{figure*}[t]
	\centering
	\includegraphics[width=0.90\textwidth]{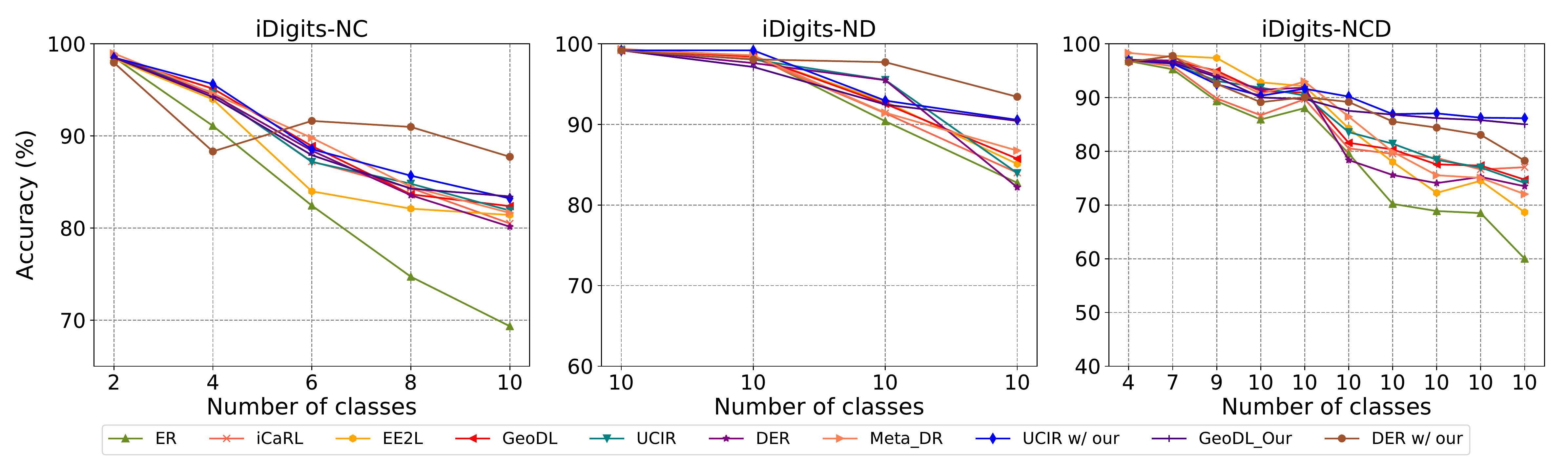}
	\caption{\textbf{Performances w.r.t sessions} on iDigits benchmark with three splits}
	\label{fig:steps_perf_idigits}
\end{figure*} 

\begin{figure*}[t]
	\centering
	\includegraphics[width=0.90\textwidth]{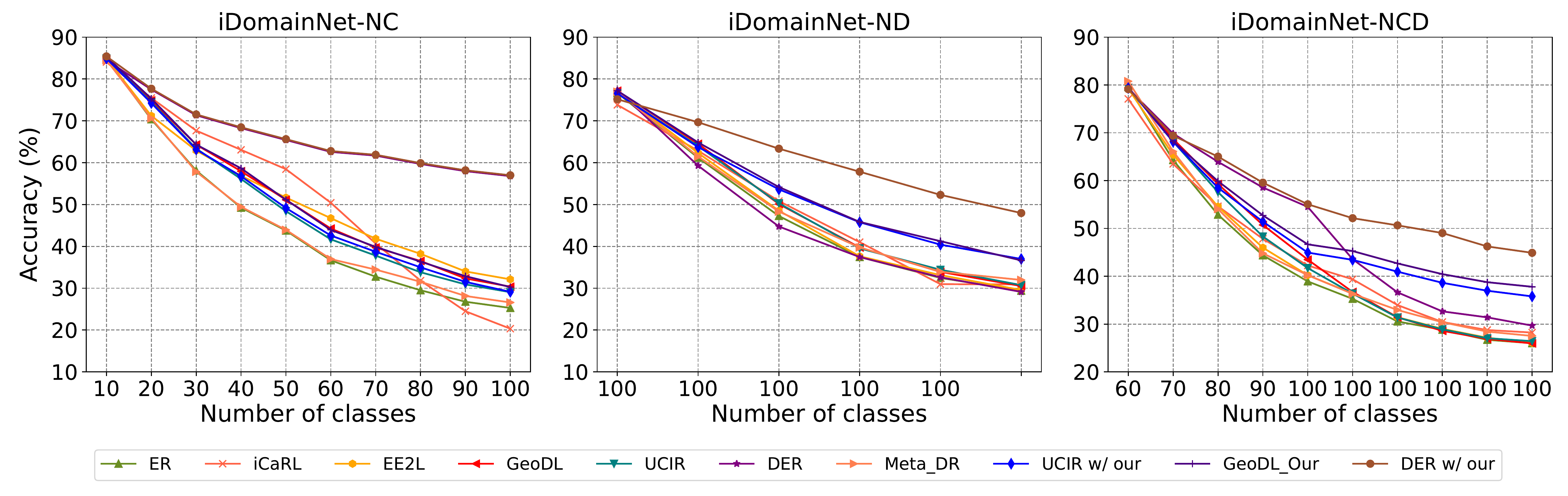}
	\caption{\textbf{Performances w.r.t sessions} on iDomainNet benchmark with three splits}
	\label{fig:steps_perf_idomainnet}
\end{figure*} 

We include more curves of performance w.r.t. sessions on iDigits and iDomainNet benchmarks with three splits, which are shown in Fig.~\ref{fig:steps_perf_idigits} and Fig.~\ref{fig:steps_perf_idomainnet}. We can see our method consistently perform better than other methods.

\section{Forgetting Metric}
We also compute the forgetting metric of different methods on the three splits of iCIFAR-20 benchmark as follows:
\begin{equation}
    F = \frac{1}{N-1}\sum_{i=2}^{N} A_{i}^{i-1} - A^{i-1}_{i-1}
\end{equation}
where $N$ is the number of total sessions. $A_{i}^{i-1}$ is the accuracy of model at session $i$ evaluated on the test set of session $i-1$. $A_{i}^{i-1} - A^{i-1}_{i-1}$ measures model's forgetting on previously observed classes and domains.
Results are presented in Tab.~\ref{tab:forget}. This shows our method has less forgetting than previous methods, which is consistent with our conclusion.

\begin{table}[t]
	\caption{Forgetting metric on iCIFAR-20.}
	\centering
		\begin{tabular}{c|ccc}
			\toprule
			Method & NC ($\%$)       & ND ($\%$) & NCD ($\%$)  \\
			\midrule
			UCIR          & -12.22 & -12.36 & -6.48 \\   
			UCIR w/ ours  & -10.66 & -10.27 & -4.48 \\
			\midrule
			GeoDL          & -11.85 & -11.33 & -4.06 \\   
			GeoDL w/ ours  & -9.12 & -8.57 & -3.73 \\
			\midrule
			DER          & -5.92 & -14.76 & -5.79 \\   
			DER w/ ours  & -5.36 & -3.44 & -1.79 \\
			\bottomrule
		\end{tabular}
	\label{tab:forget}
\end{table}

\section{More Sensitive Studies}
We examine the influence of the number of new components $m$ added at every new session, which conducted on the iCIFAR-20 ND split with 3 sessions. Session 1 has one domain for each class and sessions 2, 3 have two domains for each class. The results are shown in Tab.~\ref{tab:component}. We can see that model with small $m$ cannot perform well on this split since classes in sessions 2 and 3 have more than one domain. However, model with larger $m$ can perform closely to the oracle and is also robust to the change of number $m$. Moreover, in Table~\ref{tab:study_delta}, we show the average number of components for all the classes in the final session of the iCIFAR-20 NCD split with varying $\delta$. 	
\begin{table}[h]
	\centering
	\caption{\textbf{Sensitive Study:} Influence of $\delta$ on the number of components.}
	\resizebox{0.47\textwidth}{!}{
		\begin{tabular}{c|ccccccc}
			\toprule
			Threshold $\delta$ & 0.6     & 0.65     & 0.7     & 0.75     & 0.8 & 0.85 & 0.9        \\ \midrule
			DER w/ ours &    6.6     & 6.2     & 5.4     & 5.1      & 4.7 & 4.35 & 3.45   \\ 
			\bottomrule
	\end{tabular}}
	\label{tab:study_delta}
\end{table}
As each expansion adds 30 components, the result indicates that our reduction process can eliminate many redundant components.
The average number of components decreases as the threshold $\delta$ increases, as expected.
To measure the consistency of components and domain labels, we compute the purity of the components within each class and then average them on all the classes and sessions.
The resulting average purity is 0.962 for the iCIFAR-20 NCD split, demonstrating the efficacy of our mixture model.

We also conduct sensitive study on memory size. As shown in Tab.~\ref{tab:memsize}, the performance of the model continuously improves as the memory size increases.
Furthermore, we also evaluate the effect of randomness of memory selection. Concretely, as shown in Table~\ref{tab:randomness_memsel}, we provide results of DER w/ ours for iCIFAR-20 using five different random seeds, which shows that our random selection strategy for each class is relatively robust. That been said, we believe a better selection strategy is an interesting direction for future work.

\begin{table}[h]
	\centering
	\caption{The average performance on five random seeds.}
	\resizebox{0.4\textwidth}{!}{
		\begin{tabular}{c|ccc}
			\toprule
			& NC    &  ND &   NCD  \\ \midrule
			Avg Acc($\%$)          & 82.58$\pm$0.47 & 84.19$\pm$0.50 &   82.00$\pm$0.22 \\
			\bottomrule
	\end{tabular}}
	\label{tab:randomness_memsel}
\end{table}

\begin{table}[t]
	\caption{\textbf{Sensitive Study:} Influence of $m$ for our method on iCIFAR-20.}
	\centering
		\begin{tabular}{c|cc}
			\toprule
			Method & Final ($\%$)         & Avg ($\%$)  \\
			\midrule
			DER                  & 80.94 & 85.59 \\   
			DER w/ ours ($m=1$)  & 82.45 & 86.34 \\
			DER w/ ours ($m=5$)  & 84.39 & 87.32 \\
			DER w/ ours ($m=10$) & 84.54 & 87.40 \\
			DER w/ ours ($m=15$) & 84.08 & 87.21 \\
			\bottomrule
		\end{tabular}
	\label{tab:component}
\end{table}

\begin{table*}[t]
	\caption{\textbf{Sensitive Study:} Influence of memory size on our method for iCIFAR-20 NCD split.}
	\centering
	\resizebox{\textwidth}{!}{
			\begin{tabular}{c|cc|cc|cc|cc|cc|cc}
					\toprule
					\multicolumn{1}{c|}{\multirow{2}{*}{Memory size}} & \multicolumn{2}{c|}{$M=500$} & \multicolumn{2}{c|}{$M=1000$} & \multicolumn{2}{c|}{$M=1500$} & \multicolumn{2}{c|}{$M=2000$} & \multicolumn{2}{c|}{$M=2500$} & \multicolumn{2}{c}{$M=3000$} \\
					\multicolumn{1}{c|}{}                             & Final         & Avg          & Final         & Avg          & Final         & Avg          & Final         & Avg          & Final         & Avg          & Final         & Avg          \\ \midrule
					DER w/ ours                                       & 66.45         & 75.04        & 72.47         & 78.02        & 75.13         & 80.63        & 76.40         & 82.17        & 77.56         & 82.61        & 78.80         & 82.80    \\   
					\bottomrule
			\end{tabular}}
	\label{tab:memsize}
\end{table*}

\section{Analysis of Intra-class Imbalance Issue}

Tab.~\ref{tab:imbalance} shows the accuracy from domain 1 to domain 5 within classes  at the last session of iCIFAR-20 NCD split. As the results show, previous methods suffer from intra-class imbalance problem, which have high performance on new domains and low performance on old domains. By contrast, our method can obtain a more balanced results.

\begin{table*}[t]
	\caption{\textbf{Analysis:} Accuracies of each domains within classes in the last session of iCIFAR-20 NCD split.}
	\centering
	\resizebox{\textwidth}{!}{
		\begin{tabular}{c|c|ccccc}
			\toprule
			\multicolumn{1}{c|}{Category} & \multicolumn{1}{c|}{Method} & \multicolumn{1}{c}{Domain 1 (\%)} & \multicolumn{1}{c}{Domain 2 (\%)} & \multicolumn{1}{c}{Domain 3 (\%)} & \multicolumn{1}{c}{Domain 4 (\%)} & \multicolumn{1}{c}{Domain 5 (\%)} \\ 
			\midrule
			\multicolumn{1}{c|}{\multirow{2}{*}{reptiles}} & DER & 30.23 & 52.01 & 25.24 & 48.27 & 76.28 \\   
			\multicolumn{1}{c|}{}                 & DER w/ ours & 48.17 & 57.79 & 45.48 & 61.91 & 66.28 \\ 
			\midrule
			\multicolumn{1}{c|}{\multirow{2}{*}{medium-sized mammals}} & DER & 32.56 & 17.31 & 37.36 & 37.63 & 88.47 \\   
			\multicolumn{1}{c|}{}                 & DER w/ ours & 84.27 & 46,87 & 46.14 & 79.16 & 85.25 \\ 
			\midrule
			\multicolumn{1}{c|}{\multirow{2}{*}{large carnivores}} & DER & 33.19 & 22.00 & 71.43 & 14.18 & 92.38 \\   
			\multicolumn{1}{c|}{}                 & DER w/ ours & 69.33 & 68.59 & 85.35 & 42.74 & 85.98 \\ 
			\midrule
			\multicolumn{1}{c|}{\multirow{2}{*}{large omnivores and herbivores}} & DER & 42.41 & 66.74 & 17.65 & 45.22 & 86.11  \\   
			\multicolumn{1}{c|}{}                   & DER w/ ours & 70.23 & 64.27 & 56.48 & 84.26 & 76.32 \\ 
			\bottomrule
   	\end{tabular}}
	\label{tab:imbalance}
\end{table*}


\section{More t-SNE visualization}
\begin{figure*}[t]
	\centering
	\includegraphics[width=0.9\textwidth]{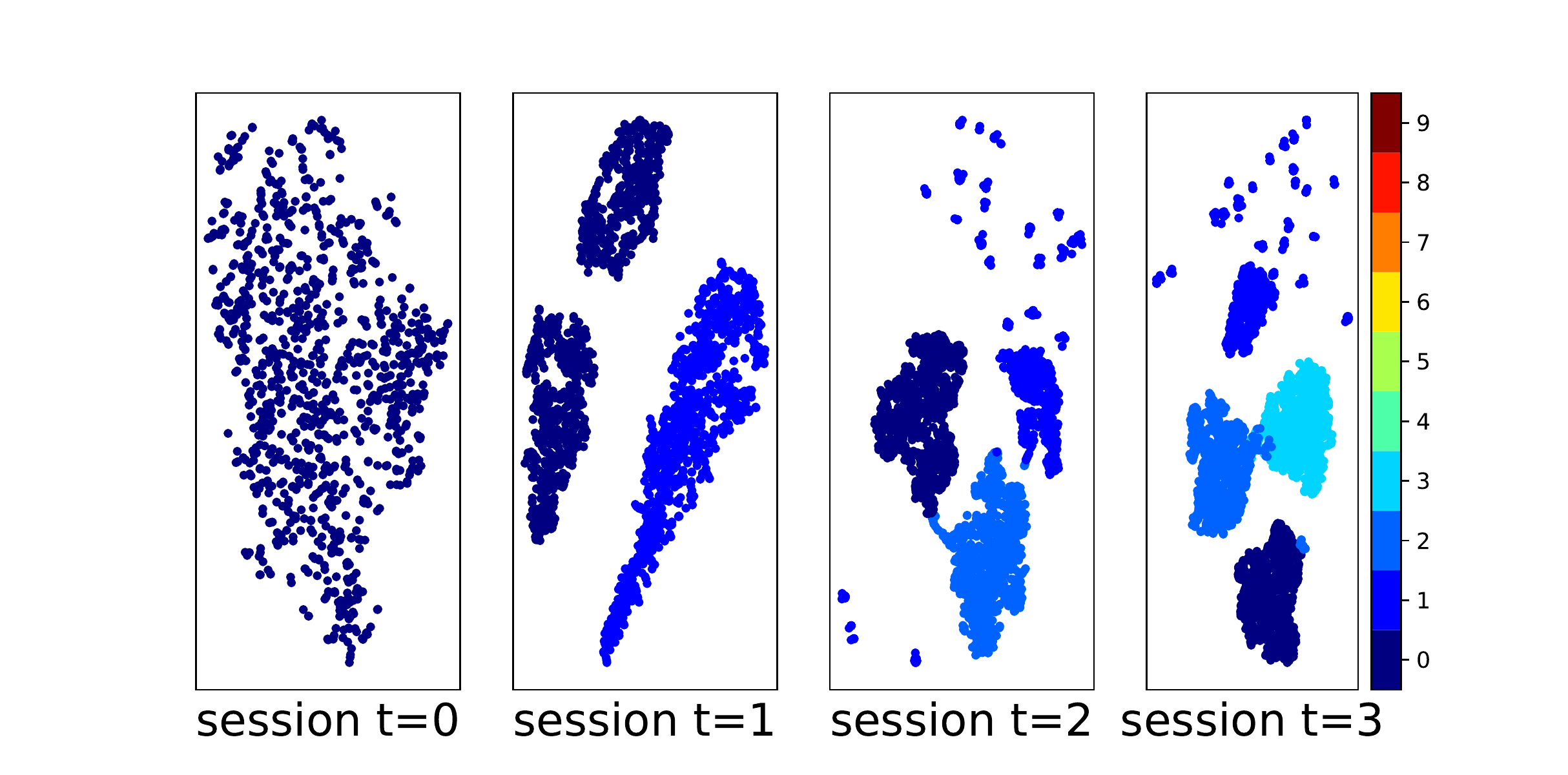}
	\caption{t-SNE visualization of digits 1.}
	\label{fig:tsne_digits1}
\end{figure*}

\begin{figure*}[t]
	\centering
	\includegraphics[width=0.9\textwidth]{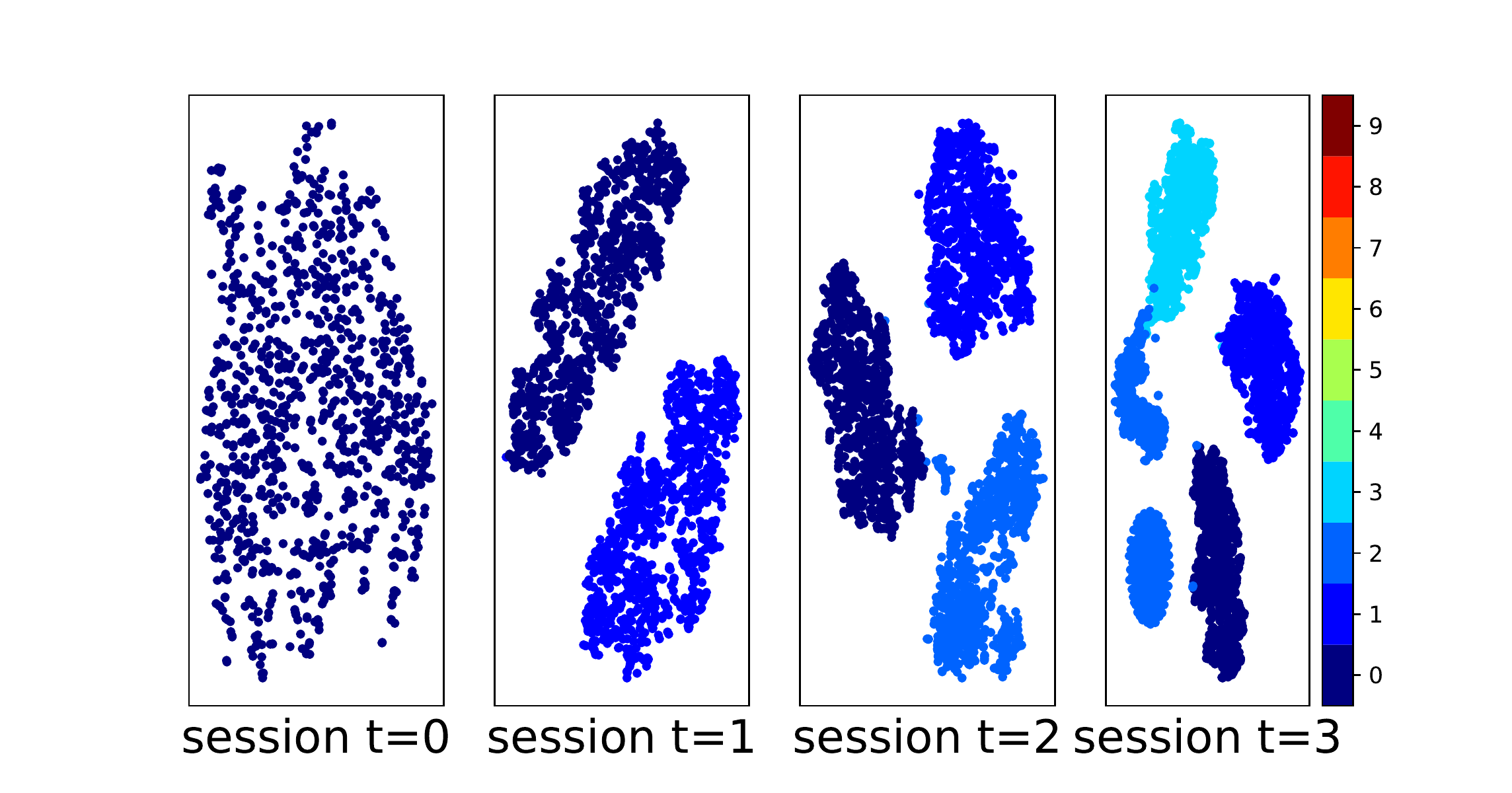}
	\caption{t-SNE visualization of digits 3.}
	\label{fig:tsne_digits3}
\end{figure*}

\begin{figure*}[t]
	\centering
	\includegraphics[width=0.9\textwidth]{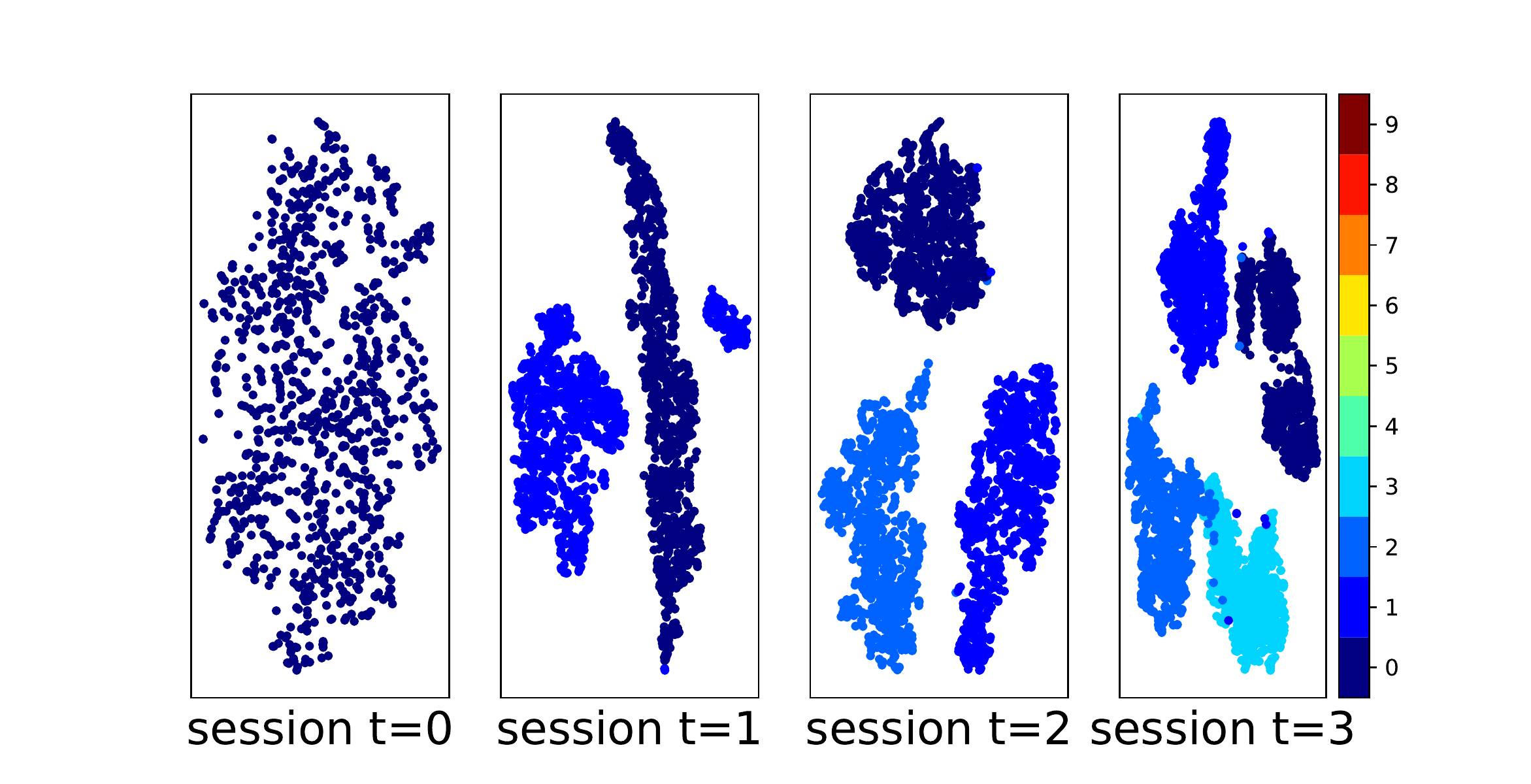}
	\caption{t-SNE visualization of digits 5.}
	\label{fig:tsne_digits5}
\end{figure*}

\begin{figure*}[t]
	\centering
	\includegraphics[width=0.9\textwidth]{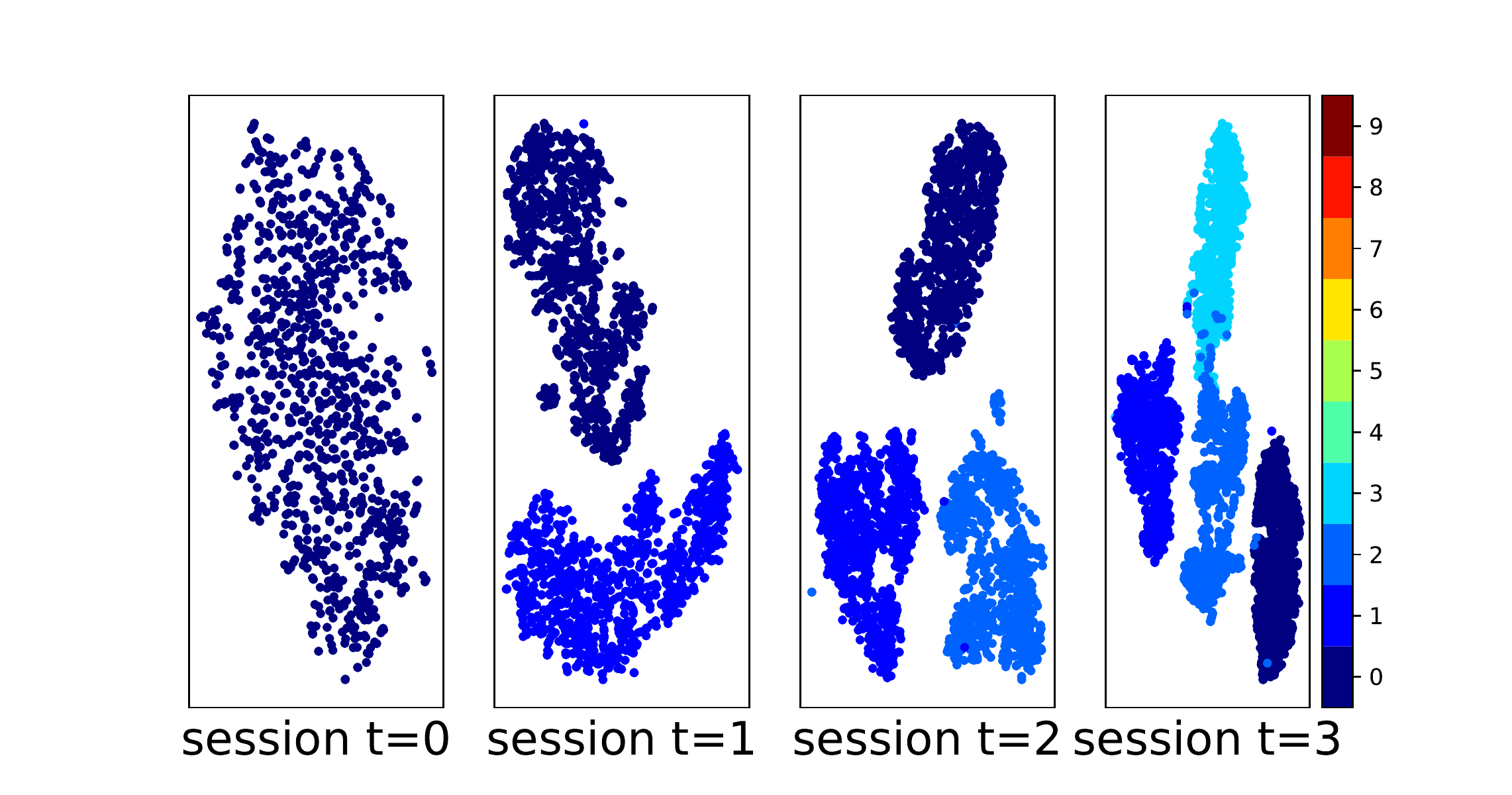}
	\caption{t-SNE visualization of digits 7.}
	\label{fig:tsne_digits7}
\end{figure*}

We provide more t-SNE visualization results of different digits in the iDigits ND split across different sessions. We can see that for most digits, shown in Fig.~\ref{fig:tsne_digits1},\ref{fig:tsne_digits3},\ref{fig:tsne_digits5},\ref{fig:tsne_digits7}, our method can inference the correct domain labels.

\end{appendices}

\end{document}